\documentclass[11pt]{article}
\usepackage{graphicx}
\IfFileExists{xurl.sty}{\usepackage{xurl}}{} %
\usepackage[numbers]{natbib}
\usepackage{parskip}
\usepackage[a4paper, total={6in,8in}]{geometry}

\usepackage[final,expansion=alltext]{microtype}

\usepackage{tikz}
\usetikzlibrary{positioning}

\usepackage[acronym,nowarn]{glossaries}
\makeglossaries
\newacronym{RCT}{RCT}{randomized controlled trial}
\newacronym{AJCC}{AJCC}{American Joint Committee on Cancer}
\newacronym{dag}{DAG}{directed acyclic graph}

\author{Wouter van Amsterdam, Pim de Jong, Joost Verhoeff,\\ Tim Leiner and Rajesh Ranganath}

\title{From algorithms to action: improving patient care requires causality}

\begin{document}
 
\maketitle

\section*{Abstract}

In cancer research there is much interest in building and validating outcome predicting outcomes to support treatment decisions.
However, because most outcome prediction models are developed and validated without regard to the causal aspects of treatment decision making, many published outcome prediction models may cause harm when used for decision making, despite being found accurate in validation studies.
Guidelines on prediction model validation and the checklist for risk model endorsement by the American Joint Committee on Cancer do not protect against prediction models that are accurate during development and validation but harmful when used for decision making.
We explain why this is the case and how to build and validate models that are useful for decision making.

\section*{Keywords}

Causal inference; oncology; tailored treatment decision making; prediction research; prognosis research

\paragraph{Introduction.}

Treatment decisions in cancer care are guided by treatment effect estimates from \glspl{RCT}.
\Glspl{RCT} estimate the \emph{average} effect of one treatment versus another in a certain population.
However, treatments may not be equally effective for every patient in a population.
Knowing the effectiveness of treatments tailored to specific patient and tumor characteristics would enable individualized treatment decisions.
Getting tailored treatment effects by averaging outcomes in different patient subgroups in \glspl{RCT} requires an infeasible number of patients to have sufficient statistical power in all relevant subgroups for all possible treatments.
Instead, we must rely on statistical modeling, potentially using observational data from non-randomized studies to further the individualization of treatment decisions.

The \gls{AJCC} recommends that researchers develop outcome prediction models in an effort to individualize treatment decisions \citep{kattan_american_2016,amin_eighth_2017}.
Outcome prediction models, sometimes called risk models or prognosis models, use patient and tumor characteristics to predict a patient outcome such as cancer recurrence or overall survival.
The assumption is that the predictions are useful for treatment decisions using rules such as ``prescribe chemotherapy only if the outcome prediction model predicts the patient has a high risk of recurrence''.
Many outcome prediction models are published every year.
Recognizing the importance of reliable predictions, the \gls{AJCC} published a checklist for outcome prediction models to ensure dependable prediction accuracy in the patient population for which the outcome prediction model was designed \citep{kattan_american_2016}.
However, accurate outcome predictions do not imply that these predictions yield good treatment decisions.
In this comment, we show that outcome prediction models rely on a fixed treatment policy which implies that outcome prediction models that were found to accurately predict outcomes in validation studies can still lead to patient harm when used to inform treatment decisions.
We then give guidance on how to evaluate whether a model has value for decision-making and how to develop models that are useful for individualized treatment decisions.

\paragraph{Accurate predictions have unknown value for decision-making.}
Individualizing treatment decisions means changing the \emph{treatment policy}.
For example, if for a specific cancer type and stage the current treatment policy is to give the same treatment to all patients, then individualizing treatment decisions means recommending treatments tailored to a patient's characteristics.
The value of an outcome prediction model is not in how well it predicts under a certain historic treatment policy, but rather what is the effect of deploying this model on treatment decisions and patient outcomes?

\begin{figure}[htpb]
	\centering
	\includegraphics[width=1\textwidth]{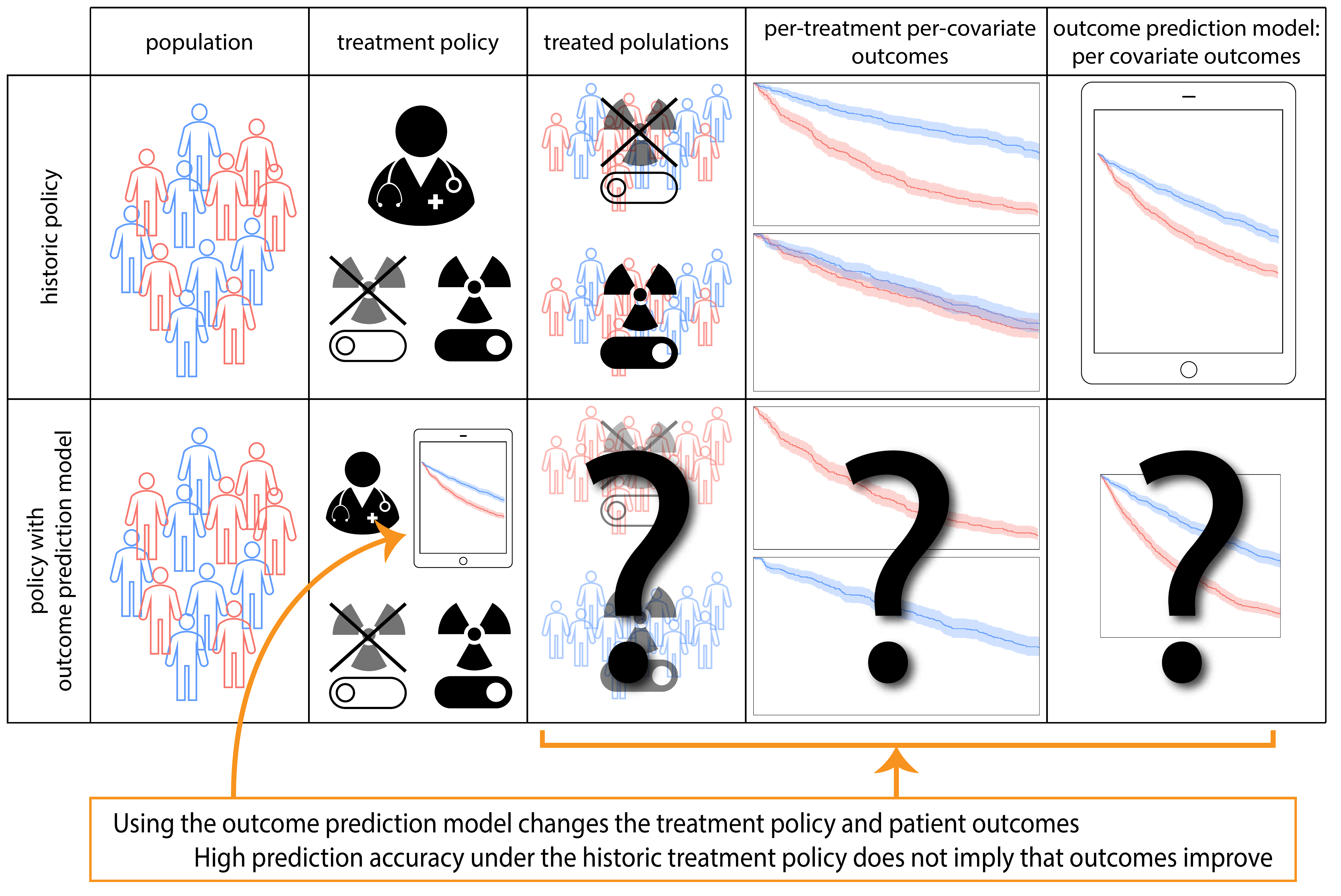}
	\caption{Illustration of the use of outcome prediction models that ignore treatment allocations in the historical data (i.e. are \textit{treatment naive}) for treatment decision making. These models change the treatment decisions and thus patient outcomes but whether this change improves patient outcomes is not determined by the prediction accuracy of the outcome prediction model.
    }
	\label{fig:txnaive}
\end{figure}

Consider an outcome prediction model that uses pre-treatment tumor characteristics to predict an outcome but ignores whatever treatment the patients may have had, i.e. \emph{treatment-naive} models, such as \citet{salazar_gene_2011,merli_simplified_2021,courtiol_deep_2019}.
Interestingly, the decision to ignore treatments in the outcome prediction model is in line with the \gls{AJCC} checklist for outcome prediction models (item 12 \citep{kattan_american_2016}).
However, these outcome prediction models can cause more harm than good when used to support treatment decisions, even when they are accurate under the historic treatment policy.
Consider for example an outcome prediction model that predicts overall survival for stage IV lung cancer patients based on the pre-treatment growth-rate of the tumor.
An accurate model would predict shorter survival for patients with faster growing tumors.
Applying this outcome prediction model, a clinician could decide to refrain from palliative radiotherapy in patients with faster growing tumors under the assumption that their life expectancy is too short to benefit from radiotherapy.
This decision based on the outcome prediction model would be unjustified and harmful, as faster growing tumors are more susceptible to radiotherapy \citep{breur_growth_1966}.
See Figure \ref{fig:txnaive} for an illustration of introducing an outcome prediction model for treatment decisions.

\paragraph{Prospective validation does not test value for decision-making.}
The gold standard for evaluating the accuracy of an outcome prediction model is \emph{prospective validation} \citep{kattan_american_2016, moons_transparent_2015}.
In a prospective validation, patient characteristics and outcomes are recorded for a new patient cohort according to a predefined protocol.
Comparing the outcome prediction model's predictions with the observed outcomes provides an estimate of how accurate the outcome prediction model is outside the cohort in which the model was developed.
The outcome prediction model from the lung cancer example above, if well-estimated, would be found accurate in a prospective validation that uses the historic treatment policy because the outcome prediction model was developed under the same historic policy.
It would then fulfill all the \gls{AJCC} checklist items but still lead to patient harm when used for treatment decisions because the differential effect of radiotherapy depending on tumor growth-rate is not accounted for in the outcome prediction model.

As an additional validation step, one may conduct a prospective validation study where the outcome prediction model is used for treatment decisions in new patients, thus changing the treatment policy.
If such a validation were carried out for the lung cancer survival outcome prediction model, the patients with fast-growing tumors would be given radiotherapy less often due to the predictions of the outcome prediction model, leading to even worse survival for these patients than before introduction of the outcome prediction model.
Introducing the outcome prediction model for decision making caused harm because under the new policy treatments are withheld from those who would have benefited most (the patients with fast-growing tumors).
However, in this validation study with model deployment the prediction model is still accurate as the model already predicted that patients with fast-growing tumors have a poor prognosis.

\paragraph{Models should improve decisions.}
The crux of the issue with outcome prediction models is that they answer the question ``What is the chance of the outcome given these patient and tumor characteristics, \emph{with the assumption that we will keep making the same treatment decisions as we always did}?''.
Similar issues exist with other kinds of outcome prediction models which make predictions using the historical treatments but without regards to the policy for how those treatments were assigned (i.e. \emph{post-decision} models such as \citet{ryu_prognostic_2014,fried_stage_2016,hippisley-cox_development_2017,liu_systematic_2022,pires_da_silva_clinical_2022}).
Post-decision outcome prediction models are also in line with the \gls{AJCC} checklist (item 12 \citep{kattan_american_2016}). To improve treatment decisions however, we need models with a foreseeable positive effect on outcomes when used in decision-making. 

\begin{figure}[!t]
	\centering
	\includegraphics[width=\textwidth]{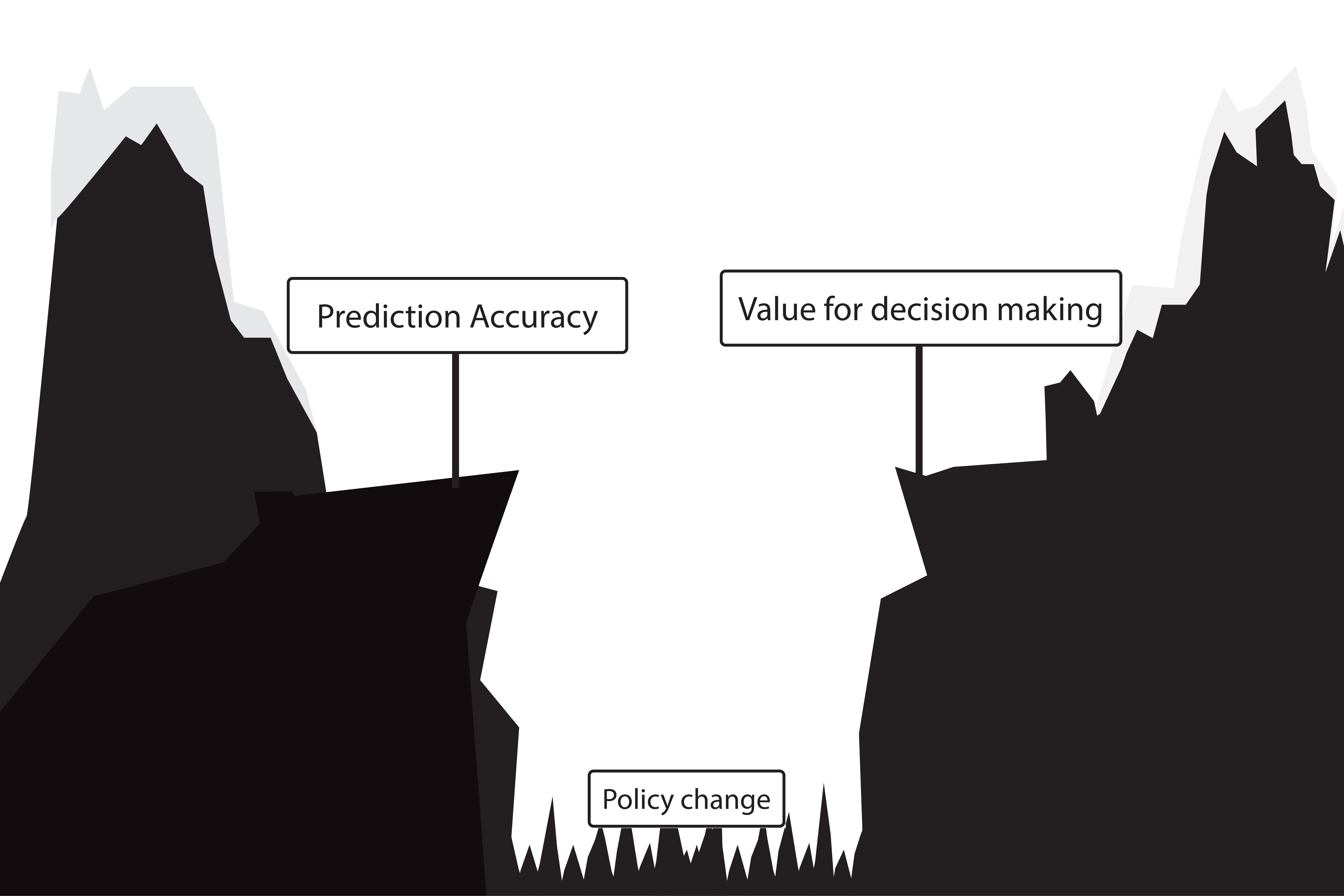}
	\caption{Illustration of the difference between outcome prediction model accuracy and its value for treatment decision making.
	Validation of an outcome prediction model following the \gls{AJCC} checklist leads to a reliable estimate of the outcome prediction model's accuracy.
	However, because the outcome prediction model relies on a fixed historic treatment policy, prediction accuracy does not imply value for decision making, as visualized with the gap.
        This gap can only be bridged with \emph{causality}.}
	\label{fig:mindthegap}
\end{figure}

Outcome prediction models assume treatment decisions follow the historical policy and thereby cannot inform us on the effect of a new policy derived from the outcome prediction model.
This reliance on the historical treatment policy leads to a fundamental gap between a prediction model's accuracy and its value for treatment decision-making in clinical practice (Figure \ref{fig:mindthegap}).
Bridging the gap from prediction accuracy to value for decision making is only possible with \emph{causality}.
Evaluating the effect of a prediction model-based treatment policy on patient outcomes requires a causal study design or causal assumptions.

\paragraph{How to validate models used for treatment decisions?}
\label{sec:validation}

The ultimate test of the effect of introducing a new treatment policy for example based on an outcome prediction model is a \emph{cluster randomized controlled trial} \citep{moons_transparent_2015, moons_prognosis_2009}.
In a cluster \gls{RCT} with outcome prediction models, some groups of clinicians are randomly selected to get access to the model while others are not.
This allows for the estimation of the effect of introducing the model on treatment decisions and patient outcomes.
For example, the cluster \gls{RCT} could demonstrate that using the model leads to fewer treatment side effects and better overall survival.
However, in the context of shared decision-making, patients may weigh the value of overall survival versus treatment discomfort differently \citep{barry_shared_2012}.
These individual preferences need to be taken into account in the cluster \gls{RCT} when calculating the value of introducing a model for decision-making.

As an alternative to cluster RCTs, the expected outcomes under a treatment policy (e.g. based on a prediction model) can be evaluated in data from a standard RCT.
This can be done by calculating the average outcome in the subgroup of patients for whom the randomized treatment assignment was concordant with the policy \citep{karmali_blood_2018}.
Multiple policies can be compared this way, for example comparing a policy based on a new prediction model with current clinical practice.
The policy with the best outcomes is preferable.
However, such an analysis does not take into account that in practice the compliance with the new treatment policy might not be perfect.
Notably, the validation steps recommended in the \gls{AJCC} checklist \citep{kattan_american_2016} provide no information on what the effect is of deploying an outcome prediction model on treatment decisions and patient outcomes.

\paragraph{Building models to individualize treatment decisions.}

Cluster RCTs are costly and time consuming.
With tools from causal inference we can improve the chance of success of models for decision making.
One way to construct a good individualized treatment policy is with models that predict the outcome under hypothetical interventions, where the intervention is the decision to give a certain treatment.
The optimal treatment policy selects the treatment that leads to the most beneficial expected outcome.

Estimating models for \emph{prediction under intervention} requires unconfoundedness, which holds when there are no unknown variables that influence both the treatment assignment and the outcome (i.e. confounders).
\Glspl{RCT} are ideal for this as unconfoundedness holds by design because the treatment assignment is random. 
However, individual \glspl{RCT} are generally too small to include many important patient and tumor characteristics in the modeling.
Observational data from regular clinical practice on the other hand are often more readily available.
If all variables that influence the treatment policy are available in a particular dataset, meaning that unconfoundedness holds, there are many approaches to prediction under intervention.
These include `conventional' statistical approaches such as regression, or machine learning approaches, for example using neural networks \citep{shalit_estimating_2017}.

\begin{figure}[htpb]
\begin{center}
	\includegraphics[width=.7\textwidth]{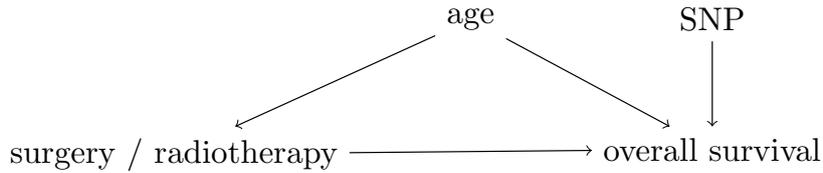}
\end{center}
\caption{Simplified Directed Acyclic Graph for the decision between surgery and radiotherapy for overall survival in lung cancer patients.
As an example, consider a hypothetical study in early-stage lung cancer where researchers investigate whether the relative effectiveness of surgery versus radiotherapy for overall survival depends on a certain single-nucleotide polymorphism (SNP).
The SNP assay was performed for the study only so this information did not affect the treatment decision.
A \gls{dag} with four variables for this study is presented in this Figure.
In this \gls{dag}, the variables \emph{age} and \emph{SNP} both have arrows to overall survival, but only \emph{age} influences the treatment decision as older patients are less likely to get surgery.
The \gls{dag} from indicates that unconfoundedness holds when \emph{age} is conditioned on in the analysis, as \emph{age} is the only confounder between the treatment and the outcome \citep{pearl_causality_2009}.
}%
\label{fig:dag}
\end{figure}

To express available background knowledge and judge whether unconfoundedness holds in observational data, researchers can use \glspl{dag} \citep{pearl_causality_2009}.
\Glspl{dag} depict variables (such as treatment, outcome and confounders) and causal dependencies between the variables as \emph{arrows} that point from a cause variable to an effect variable, see Figure \ref{fig:dag} for an example.
Using tools from \emph{causal inference}, the \gls{dag} determines whether a prediction under intervention model can be estimated and if so what confounders need to be accounted for \citep{pearl_causality_2009}.

In some cases not all confounders are available.
In this setting with unobserved confounding standard methods based on confounder adjustment cannot be used, but sometimes prediction under intervention models may be estimated using specialized methods.
Two examples are methods based on proxy-variables of unmeasured confounders \citep{miao_identifying_2018,van_amsterdam_individual_2022} and instrumental variable methods \citep{wald_fitting_1940} and their machine learning variants \citep{hartford2017deep, puli2020general}.
These methods rely on assumptions that may not hold perfectly in reality, so figuratively speaking they might reduce the gap between model accuracy and treatment policy value, but not close the gap entirely.
\Glspl{dag} encode \emph{assumptions} about the data which may not hold perfectly in practice.
The effects of potential violations of these assumptions may be estimated using \emph{sensitivity analyses} \citep{greenland_basic_1996}.

A special case for prediction under intervention is the untreated risk, which is the hypothetical outcome under no treatment (or some baseline treatment) and would be observed in the control group of an \gls{RCT}.  For instance, when deciding to give adjuvant therapy after breast cancer surgery, the untreated risk of recurrence is the risk of recurrence when no adjuvant therapy would be given \citep{candido_dos_reis_updated_2017}.
Knowing the untreated risk is valuable when considering giving no further treatment, and as a baseline to compare other potential treatments against.
Although estimating the untreated risk requires unconfoundedness, in some cases it may be estimated quite accurately even from confounded data using specialized methods \citep{van_amsterdam_conditional_2023}.

Because \glspl{RCT} randomly assign patients to interventions, models for prediction under intervention can be validated in \glspl{RCT} with standard prediction validation approaches \citep{moons_transparent_2015}.
For shared decision-making, prediction under intervention of different treatment options allows the patient to make their own judgment on how to weigh e.g. expected overall survival with expected treatment discomfort.
Whereas individual \glspl{RCT} randomize a patient to a certain treatment, cluster \glspl{RCT} randomize clinicians' access to a model for decision support.
Thereby individual treatment decisions may still be confounded in cluster \glspl{RCT} meaning that cluster \glspl{RCT} cannot validate predictions from \emph{prediction under intervention}-models directly.
Both policy evaluation with cluster RCTs and prediction-under-intervention validation in standard RCTs are also possible in observational data but require unconfoundedness and thereby sensitivity analyses for potentially omitted confounders \citep{keoghPredictionInterventionsEvaluation2024, greenland_basic_1996}.

\section*{Discussion}

In line with \acrlong{AJCC} recommendations \citep{kattan_american_2016,amin_eighth_2017} many researchers develop outcome prediction models to individualize treatment decisions.
The \gls{AJCC} checklist provides important guidelines for outcome prediction model development and validation, such as clearly defining the patient population, predictor variables and prediction time-point, in addition to validation in external datasets.
These items improve the dependability of outcome prediction models for predicting outcomes in the intended patient population if there are no changes in the treatment policy \citep{kattan_american_2016}.
However, not changing the treatment policy directly contradicts the intended purpose of these models.
Outcome prediction models that satisfy all the criteria in the checklist still have unknown clinical utility because high prediction accuracy in prospective validation studies does not imply value for treatment decision-making in clinical practice \cite{vanamsterdamWhenAccuratePrediction2024a}.
Because the gap between outcome prediction model accuracy and value for decision-making is due to causal issues, it is not resolved by larger datasets, more flexible prediction algorithms (e.g. machine learning) or even by prospective validation with model deployment.
In contrast, we explained how models for prediction under intervention are useful for decision-making and how to validate any model used in decision-making.

The gap between outcome prediction model accuracy and value for decision-making is due to causal issues, but it is different from the standard ``correlation does not imply causation''.
In the standard ``correlation is not causation'' setting, all variables (treatment, outcome, patient/tumor characteristics) are already present in the historical data, whereas in this case, the output of the outcome prediction model cannot be a cause of the outcome.
This is because the outcome prediction model is not a variable in historical data, but a shift in policy that changes the distribution of the treatment.

It was noted before that cluster \glspl{RCT} are the ultimate test for the impact of a new prediction model on clinical practice due to issues related to compliance with treatment recommendations \citep{moons_prognosis_2009}.
We show that because of the gap between prediction accuracy and value for treatment decision-making, many accurate outcome prediction models will fail to demonstrate value in cluster \glspl{RCT}.
Also, cluster RCTs measure the effect of a new policy on average outcomes but do not directly measure whether a model accurately predicts the outcome under intervening to give a certain treatment, for this individually randomized data are most valuable.
For shared decision-making accurate predictions-under-intervention may be most important.
Two patients with the same predicted outcomes may make a different treatment decision because each patient has their own values and preferences.
In a cluster RCT, these individual values need to be accounted for when evaluating a treatment policy, for example by eliciting the values and incorporating them in the analysis when weighing for example overall survival and treatment discomfort.

Previous work underlined the value of \emph{prediction-under-intervention} models (sometimes referred to as \emph{counterfactual prediction}) for supporting treatment decisions \citep{prosperi_causal_2020,van_geloven_prediction_2020}.
Our comment highlights the potential harm of current common practice where outcome prediction models are deployed for decision making based on prediction accuracy alone, further emphasizing the relevance of prediction-under-intervention.
In addition, we note how models may be validated for decision support for example with cluster \glspl{RCT}.

\begin{figure}[!t]
	\centering
	\includegraphics[width=\textwidth]{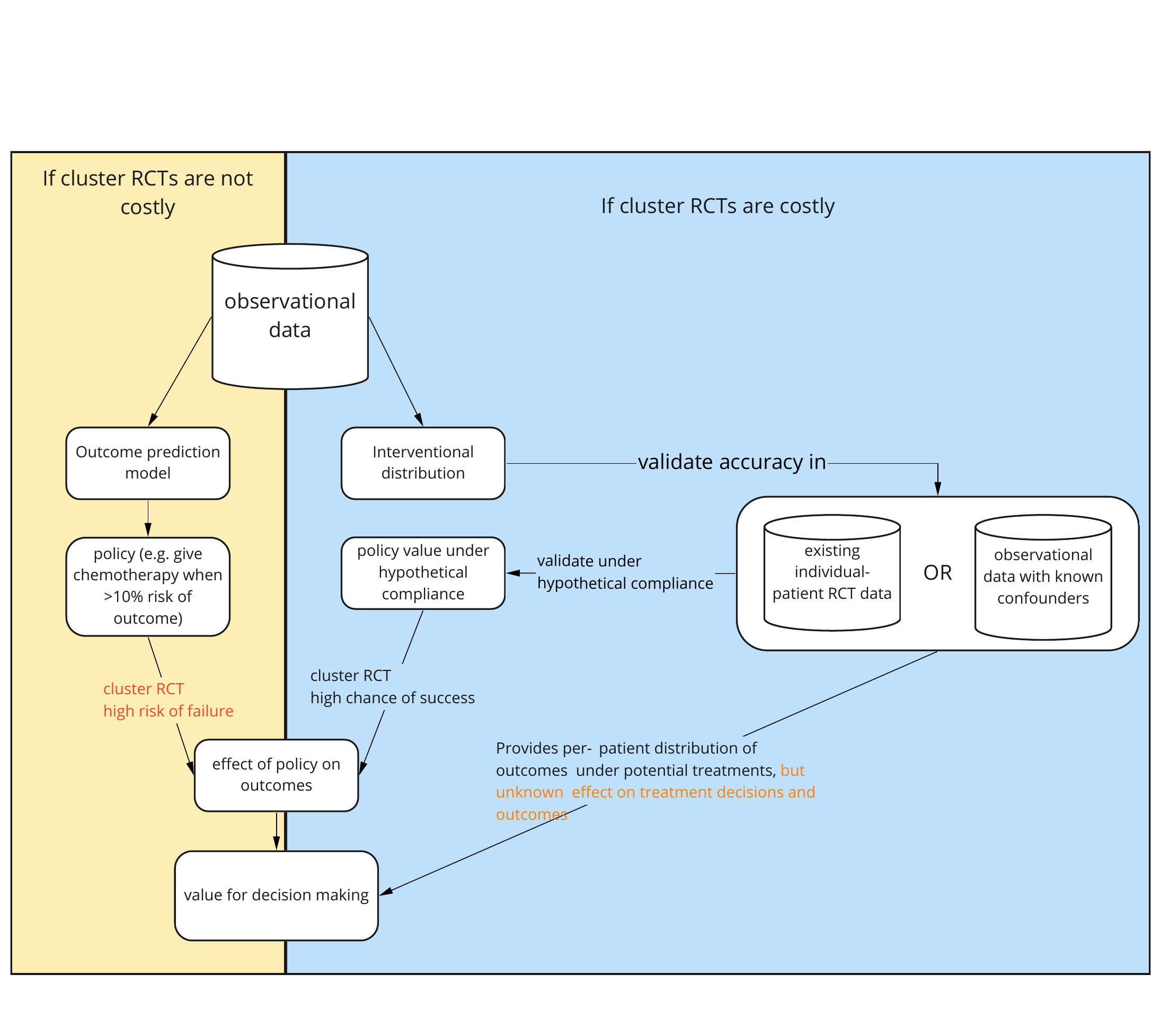}
	\caption{Flowchart of what to do depending on the costliness of cluster randomized controlled trials.
		Costliness of cluster \glspl{RCT} should be taken broadly, including time, money and ethical considerations.
	}
	\label{fig:whattodo}
\end{figure}

Building models for prediction under intervention is harder than developing outcome prediction models due to the extra requirement of unconfoundedness, which involves formalizing assumptions about confounders for example with \glspl{dag}, gathering data on all confounders, often more complex statistical estimation, and sensitivity analyses.
When the cost to do a cluster \gls{RCT} is low, it may suffice to build outcome prediction models in line with the \gls{AJCC} checklist and test them in cluster \glspl{RCT} before model deployment.
As illustrated in Figure~\ref{fig:whattodo}, when cluster \glspl{RCT} are costly, impractical or unethical, models that predict under interventions are preferable as they have
foreseeable effects when used for treatment decision-making.

There is a classical distinction between treatment effect estimation and prediction that amounts to ``treatment effect estimation is causal (and thus requires \glspl{RCT})'' but ``prediction is not causal''.
When it comes to individualizing treatment decisions with prediction models, this distinction is unhelpful and confusing as the goal is to predict what would happen under different interventions.
Selecting the best treatment for a patient is a causal question and requires causal answers.

\break

\printglossary

{
	\small
	\bibliographystyle{unsrtnat}
	\bibliography{comment}
}

\pagebreak

\section*{Declarations}

\subsection*{Ethics approval and consent to participate}
Not applicable

\subsection*{Consent for publication}
All authors consent with publication

\subsection*{Availability of data and materials}
Not applicable

\subsection*{Competing interests}
The authors declare no competing interests as defined by BMC, or other interests that might be perceived to influence the results and/or discussion reported in this paper.

\subsection*{Funding}
There was no specific funding for this comment.

\subsection*{Authors' contributions}
Conceptualization: WA, PJ, JV, TL, RR. Writing of draft manuscript: WA and RR. Editing and revision: all authors

\subsection*{Acknowledgements}

Drs. Lidia Barberio, Director of ``Longkanker Nederland'' (the Dutch patient association for lung cancer) provided feedback on this comment. Her input broadened the scope of this work making it more relevant for patients. Specifically, we added more emphasis on the importance of including the values of the patient in treatment decision-making.

\textbf{\hfill\break
}

\end{document}